\documentclass[10pt,twocolumn,letterpaper]{article}

\usepackage[pagenumbers]{wacv} % To force page numbers, e.g. for an arXiv version

\usepackage{graphicx}
\usepackage{amsmath}
\usepackage{amssymb}
\usepackage{booktabs}

\usepackage{multirow}
\usepackage[linesnumbered,ruled,vlined]{algorithm2e}
\usepackage{algcompatible}
\graphicspath{ {img/} }
\usepackage{caption}
\usepackage{subcaption}
\usepackage{url}
\usepackage{bbding}
\usepackage{tikz}

\usepackage[pagebackref,breaklinks,colorlinks]{hyperref}

\usepackage[capitalize]{cleveref}
\crefname{section}{Sec.}{Secs.}
\Crefname{section}{Section}{Sections}
\Crefname{table}{Table}{Tables}
\crefname{table}{Tab.}{Tabs.}

\def\wacvPaperID{7} % *** Enter the WACV Paper ID here
\def\confName{WACV}
\def\confYear{2024}

\newcommand{\OurName}{SeaDSC}

\makeatletter
\def\@fnsymbol#1{\ensuremath{\ifcase#1\or \dagger\or \ddagger\or
\mathsection\or \mathparagraph\or \|\or **\or \dagger\dagger
\or \ddagger\ddagger \else\@ctrerr\fi}}
\makeatother

\usepackage[accsupp]{axessibility}  % Improves PDF readability for those with disabilities.

\begin{document}
\pdfoutput=1
% A video-based unsupervised technique for detecting dynamic changes in unmanned surface vehicles.
\title{\OurName: A video-based unsupervised method for dynamic scene change detection in unmanned surface vehicles}

% \thanks{Corresponding author (linhtk.dhbk@gmail.com).}
\author{Linh Trinh, Ali Anwar, Siegfried Mercelis\\
{Faculty of Applied Engineering, IDLab, University of Antwerp-imec, Belgium} \\
{\tt\small\{linh.trinh, ali.anwar, siegfried.mercelis\}@uantwerpen.be}\\
}

\maketitle

%%%%%%%%% ABSTRACT
\begin{abstract}
    % The research on maritime vision has been fast expanding in recent times. There are a wide area studies on computer vision application for Unmanned Surface Vehicles (USVs). The utilization of multimodal sensors, including camera, radar, and lidar, has led to extensive research in various computer vision tasks such as object detection, segmentation, object tracking, and motion planning. The increasing importance of video analysis, indexing, browsing, summarization, compression, and retrieval systems is driven by the rapid expansion of video data collected for maritime vision.   Video scene change detection is an initial and crucial stage in applications related to pictorial information retrieval and scene understanding. Nevertheless, the research topic concerning maritime remains uncommon.
    Recently, there has been an upsurge in the research on maritime vision, where a lot of works are influenced by the application of computer vision for Unmanned Surface Vehicles (USVs). Various sensor modalities such as camera, radar, and lidar have been used to perform tasks such as object detection, segmentation, object tracking, and motion planning. A large subset of this research is focused on the video analysis, since most of the current vessel fleets contain the camera's onboard for various surveillance tasks. Due to the vast abundance of the video data, video scene change detection is an initial and crucial stage for scene understanding of USVs. This paper outlines our approach to detect dynamic scene changes in USVs. To the best of our understanding, this work represents the first investigation of scene change detection in the maritime vision application. Our objective is to identify significant changes in the dynamic scenes of maritime video data, particularly those scenes that exhibit a high degree of resemblance. In our system for dynamic scene change detection, we propose completely unsupervised learning method. In contrast to earlier studies, we utilize a modified cutting-edge generative picture model called VQ-VAE-2 to train on multiple marine datasets, aiming to enhance the feature extraction. Next, we introduce our innovative similarity scoring technique for directly calculating the level of similarity in a sequence of consecutive frames by utilizing grid calculation on retrieved features. The experiments were conducted using a nautical video dataset called RoboWhaler to showcase the efficient performance of our technique.
    % The implementation of our approach can be accessed by the public through this repository (to be provided). 
\end{abstract}

%%%%%%%%% BODY TEXT
\section{Introduction}
\label{sec:intro}
Approximately 80\% of global trade is conducted through maritime transportation \cite{survey_general}. As the use of sea based transport continues to rise, there is a corresponding increase in incidents such as pirate attacks, trafficking of illegal substances, illegal immigration and fishing, terrorist attacks in port areas, and collisions between marine vehicles, particularly in inland waters, coastal shipping and near the ports. To tackle these challenges, numerous studies have introduced the application of machine learning and deep learning to address various computer vision problems.
Extensive research has been conducted on the application of computer vision in USVs to advance the development of autonomous shipping and vessel systems. Equipped with advanced multimodal sensors such as cameras, radar, and lidar \etc, much research has been conducted on various computer vision tasks to assure the practical implementation of USVs in real-world scenarios. Several recent literature reviews have provided a systematic overview of various techniques and algorithms used in maritime computer vision. These include maritime object detection, object tracking, segmentation \cite{survey_maritimedetection@Iwin, survey_maritimevision@Zhang}, deep learning for maritime vision \cite{survey_maritimevision_dl@Qiao,survey_maritimevision@Qiao}, and comprehensive state-of-the-art (SOTA) maritime datasets for maritime perception \cite{survey_maritimevision_dataset@Su}.
With the increasing expansion of video data collected for marine vision, the significance of video analysis, indexing, browsing, summarization, compression, and retrieval systems has grown considerably \cite{survey_av@Badr,survey_av@Grigorescu}. The identification of scene changes in videos is the initial and crucial stage in applications related to visual information retrieval and scene understanding. Nevertheless, this research issue in the field of maritime remains uncommon.
In this paper, we present our method for dynamic scene change detection for USVs. To the best of our knowledge, this study represents the inaugural investigation of this problem into the application of maritime vision. Our objective is to identify significant changes in the dynamic scenes of maritime video data, particularly those scenes that exhibit a high degree of resemblance. Determining dynamic scene change on the maritime video will bring many benefits such that help to categorize the potential meaningful action, remove amount of redundant and unnecessary scenes or frames in the future. In this regard, the previous works can be categorized into either a supervised learning strategy or an unsupervised learning technique. Supervised learning methods \cite{dsc3@Peng,dsc4,dsc5@Feichtenhofer,dsc6@Aalok,dsc7@Liang} typically necessitate a substantial amount of data annotation for a preset set of dynamic scenes. These methods then proceed to train a model to identify dynamic scenes in videos. However, data annotation is consistently expensive, and in the emerging field of maritime vision, annotation for this specific purpose is currently unavailable.
On the other side, the unsupervised learning approach does not necessitate annotation for training the model. Alternatively, the methodologies described in  \cite{dsc_1@Salih,dsc_2@Rascioni,dsc_8@Dorfeshan,dsc_9@Rayatifard,dsc_10@Shukla} utilize a representative method that calculates the similarity between every consecutive frame in a video. This calculation is based on low-level features retrieved from the frames, such as color, histogram, gradient, and so on. At a later step, an aggregation algorithm combines all the obtained scores to determine the degree of similarity in scene changes. However, this method could result in inaccuracies since the low-level features may not include sufficient information and may contain noise. In addition, doing calculations on every pair of successive frames in the video segment results in a significant increase in the computational cost.  
Our novel approach aims to optimize the computational cost while simultaneously improving the accuracy of scene change detection for maritime applications. Initially, we utilize a recent SOTA generative deep learning model, to construct a feature extraction method that improves low-level features beyond what traditional feature extraction can do. Here, we utilize the VQ-VAE-2 model as a basis and make a minor modifications for more lightweight. We then train our model using a collection of maritime datasets in a reconstruction task, without the need for any annotations. During the inference stage, we calculate the similarity magnitude for each sliding window of the video by projecting the features onto the skipped frames of the window. In addition, we calculate the similarity score for a pair of skipping frames using cell calculation of a grid feature. This approach helps to reduce computational cost while maintaining high accuracy in the inference process.  

Our main contributions are summarized below:
\begin{itemize}
    \item We introduce our framework for detecting dynamic scene changes in a video sequence captured by USVs. Our approach utilizes successive frames and relies on unsupervised learning.
    \item We adapt a SOTA image generating VQ-VAE-2 model \cite{vqvae2} to train it on a marine dataset in order to extract sophisticated embedding features.
    \item As the heart of the component, we propose a novel feature extraction-based scheme for calculating the convulsion score of a segment on its own.
    \item We use multiple raw datasets in maritime computer vision to demonstrate the effectiveness of our approach and promising performance.
    % \item We present a new and efficient approach for calculating the convulsion score of a pair of frames using their matching feature grid.
    % \item We employ multiple raw datasets in maritime computer vision to show the effectiveness of our approach and demonstrate the favorable and encouraging performance.
\end{itemize}
In the following sections, we present existing work in Section \ref{sec:rel}, then we discuss in the detail of our methodology on the Section \ref{sec:method}, and show the results of our experiment on the Section \ref{sec:exp}. Finally, the Section \ref{sec:conclude} conclude our works.
%---------------------------------------
\section{Related works}\label{sec:rel}
\subsection{Maritime vision studies}
The field of maritime vision has been extensively researched. Several literature reviews have methodically and fully compiled and presented the most recent studies in this field.   Qiao Qiao \etal \cite{survey_maritimevision_dl@Qiao,survey_maritimevision@Qiao} introduced an extensive deep learning approach for the application of USVs in maritime environments. A comprehensive approach utilizing deep learning is proposed to address the challenges associated with USVs. This approach encompasses several tasks and methodologies including environment perception, state estimation, and path planning. These tasks are effectively tackled through the application of supervised, unsupervised, and reinforcement learning methods. 
Zhang \etal \cite{survey_maritimevision@Zhang} conducted a thorough investigation and extensive comparison of object detection approaches, with a particular emphasis on the maritime domain. Their study specifically examined deep learning models, including CNN-based methods and YOLO-based methods, from 2012 to 2021. Iwin \etal \cite{survey_maritimedetection@Iwin} conducted a comprehensive analysis of the existing research on deep learning techniques for ship object detection and recognition. 
In addition to the literature studies on USVs approaches, a literature study is also being undertaken on maritime vision datasets. Su \etal \cite{survey_maritimevision_dataset@Su} provided a comprehensive overview of publicly available maritime datasets for the purpose of maritime perception. A total of 15 state-of-the-art public datasets are systematically presented, which consist of multi-modal sensors including cameras, lidar, and radar sensors for data collecting.  

\begin{figure*}[ht]
    \centering
    \includegraphics[width=0.9\textwidth]{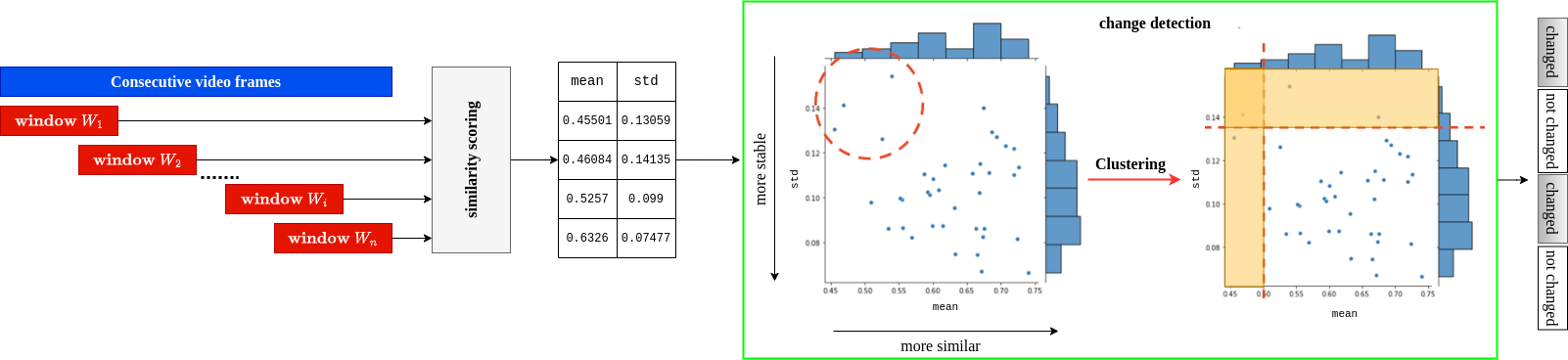}
    \caption{Our proposed {\OurName} framework for scene change detection for USVs.}
    \label{fig:fw}
\end{figure*}
\subsection{Scene change detection methods}
Kowdle \etal \cite{dsc_eccv2012@Kowdle} introduced a technique for detecting dynamic scene changes in videos, which can be applied to video analysis tasks like video summarization. The method suggests a computational approach for determining the similarity between each pair of frames within a sliding window. This study evaluated the model's performance on the movie dataset \cite{dsc_dataset@Holiwood} using precision, recall, and F1-score measures. This work calculates the similarity between consecutive pairs of frames in a sliding window, which results in increased processing costs. Furthermore, the utilization of optical flow in the method results in features that are of low quality, imprecise, and affected by noise.  
Salih \etal \cite{dsc_1@Salih} present a technique for detecting dynamic scene changes in video coding. The main objective in this context is to achieve video compression by removing the temporal redundancy between consecutive frames. The four matching methods available are: Absolute Frame Difference (AFD), Mean Absolute Frame Differences (MAFD), Mean Histogram Absolute Frame Difference (MHAFD), and Maximum Gradient Value (MGV). However, the method mainly depended on handcrafted features such as color, histogram, or image gradient, which are still limited and may contain a significant amount of noise. Furthermore, the calculation of each successive pair of frames incurs a significant computational expense.  
Rascioni \etal \cite{dsc_2@Rascioni} introduced their technique for detecting dynamic scene changes in H264 encoded video. This study measure dissimilarity by analyzing different features, including pixel-level comparison, global histogram, block-based histogram, and motion-based histogram of video sequences. The objective is to detect scene changes in movies. The method primarily depended on domain-specific information for feature extraction and threshold determination, resulting in inaccuracies and unreliability for real-world applications. 
Peng \etal \cite{dsc3@Peng} proposed a method for dynamic scene identification using a sequence model called LSTM. This model aggregates the trajectory of information retrieved from CNNs to categorize scenes. 
Feichtenhofer \etal \cite{dsc4,dsc5@Feichtenhofer} introduced a method for dynamic scene detection. They utilized a feature extraction process that specifically targeted spatial and spatiotemporal information to identify dynamic scenes in subsequent stages. The proposed model relies on the ResNet backbone and requires annotated data for training, which may not always be available in various domains. 
Aalok \etal \cite{dsc6@Aalok} presented a technique for extracting features from Convolutional Neural Networks (CNNs). These features are then combined into a high-dimensional vector and fed into a Support Vector Machine (SVM) classifier to determine the dynamic scene.  However, this strategy also necessitates a substantial amount of annotated data in order to train the model. 
Liang \etal \cite{dsc7@Liang} present a system with multiple stages to train an algorithm for detecting scene changes in videos. 
Dorfeshan \etal \cite{dsc_8@Dorfeshan} presented a technique for detecting dynamic scene changes by calculating the similarity between video frames' histograms. 
Rayatifard \etal \cite{dsc_9@Rayatifard} addressed the problem of dynamic scene recognition for video segmentation. They specifically focus on HEVC/H.265 video streams, where the similarity between consecutive frames is computed using the compressed bitstream signal. This similarity is then used to identify scene changes. 
Shukla \etal \cite{dsc_10@Shukla} introduced their method for detecting scene changes in videos. Their approach involves using histograms, binary search, and linear interpolation to filter out comparable frames in the video. 

The regular metric for evaluating a (dynamic) scene change detection are recall, precision, and F1-score as all above studies for scene change detection.

%-----------------------------------
\section{Method} \label{sec:method}
Our goal is to determine whether a changed part of a video has been presented. The target scene has changed significantly and continuously over a long period of time. A static scene, on the other hand, is one that remains motionless or unchanged for an extended period of time. For example, USVs typically travel in a straight line along the beach for extended periods of time without encountering other USVs or changing their surroundings. Other instances where the camera is obstructed or another USV blocks the ego USV for an extended period of time could be considered unchanged scenes.

The framework we propose is seen in Figure \ref{fig:fw}. Within our architecture, we initially partition a given video into windows of equal duration. By utilizing a stride that is less than the length of the window, it is possible to overlap the window. Subsequently, each window is subjected to a similarity scoring method in order to determine the degree of similarity inside that window. A higher score, in essence, indicates that the entire frame in the video has a similarity to each other, implying that this window could be considered a potential instance of no scene change. The similarity scoring module will generate the mean and standard deviation of the similarity score, which is used as input for the subsequent module. Furthermore, a higher mean value indicates a greater similarity across the frames inside the window. Moreover, a smaller standard deviation value indicates that the similarity scores inside the window are more consistent and have less variance. The change detection module utilizes this input from all video windows to carry out a clustering operation using the table of mean and standard deviation values and determines if a window belongs to a changed or not changed class.

\begin{figure}[ht]
    \centering
    \includegraphics[width=0.45\textwidth]{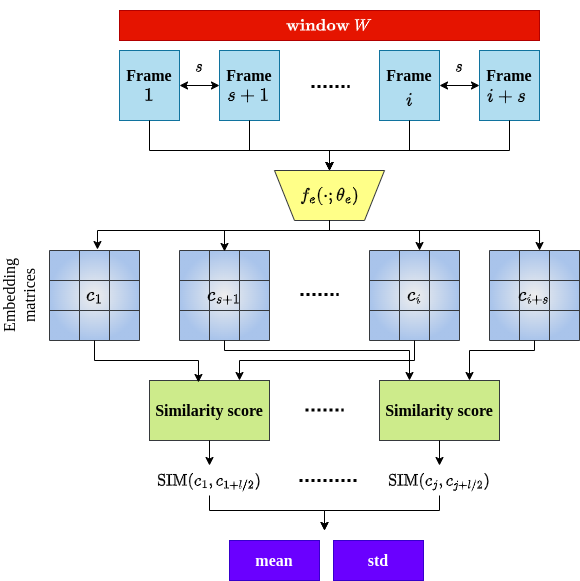}
    \caption{Similarity scoring calculation component of {\OurName}.}
    \label{fig:sim_score}
\end{figure}

The Figure \ref{fig:sim_score} displays the detailed design of our similarity scoring module. This module is specifically intended to enhance computational efficiency while effectively improving change detection performance. In order to reduce computational load, we implement a skip frame strategy by selecting only one frame out of every $s$ consecutive frames for further processing, thus avoiding the need to compute across all frames of the window. In order to circumvent the need for direct computation from the original images, we propose the utilization of a projection function $f_e(\cdot;\theta_e)$ to map each image onto its corresponding feature representation, specifically the embedding matrices ($\theta$ is the parameters of the projection model). Computation at the feature level has several advantages, as the feature level contains more rich information and is smaller in size compared to the high-dimensional pixel matrix of the original image. The specifics of our projection methodology are outlined in the subsection \ref{sub:proj_model}.
Once we have obtained the embedding matrices that represent our original images, we proceed to calculate the similarity score. Our hypothesis is that if the window undergoes a change in scenery, then the pair of frames at the start of the window will be different from the frame in the middle of the window. We calculate the similarity score between each pair of frames, specifically the $i$-th frame and the $(i+l/2)$-th frame within the window, rather than comparing consecutive frames.
% Instead of comparing consecutive frames, we calculate the similarity score between each pair of frames, specifically the $i$-th frame and the frame $(i+l/2)$-th within the window.
The specifics of how the similarity score is calculated are outlined in subsection \ref{sub:sim_score}. 

Once the similarity score calculation is complete, we determine the mean and standard deviation values as a pair to represent the similarity score of the window. These values are used in the input for the change detection component. 

\subsection{Projection model}\label{sub:proj_model}
We utilize the VQ-VAE-2 algorithm \cite{vqvae2} in our projection model $f_e(\cdot;\theta_e)$. Compared to previous generative models, VQ-VAE-2 has superior lossy compression capabilities, allowing for training models on high-resolution photos. The fundamental concept behind VQ-VAE-2 involves training many hierarchical layers, where an input picture is compressed into quantized latent maps of varying sizes at each level.
% We employ VQ-VAE-2 \cite{vqvae2} for our projection model. Compare with other generative models, VQ-VAE-2 provided the great lossy compression which enable the training model on the high resolution images. The basic idea of VQ-VAE-2 is to train mutiple hierarchical level which an input image will be compressed onto quantized latent maps with different size at each hierarchical levels.
Vector quantization is a process that involves an encoder, which maps inputs to a series of discrete latent vectors, and a decoder, which reconstructs the inputs using these discrete vectors. To be more precise, when provided with an image $\mathrm{x}$, the encoder will apply a non-linear transformation to produce a vector $E(\mathrm{x})$. The vector is subsequently quantized by comparing its distance to a set of predefined vectors in the codebook list, resulting in the selection of a certain codebook $\mathrm{e}$. Similar to the previous work using VQ-VAE \cite{trinh2022vqc}, the codebook for vector quantization is initialized using a uniform distribution as in the equation (\ref{eq:codebook}). 
% Vector quantization is conceptualized as a comprising of an encoder that maps inputs onto a sequence of discrete latent vectors, and the decoder that reconstructs the inputs from these discrete vectors. More formally, given an image $\mathrm{x}$, the encoder will non-linear transform it into an a vector $E(\mathrm{x})$. This vector then quantized based on its distance to a predefined vectors in the codebook list to a selected codebook $\mathrm{e}$. Following is the initialization of the uniform distribution-based codebook for vector quantization:
\begin{equation} \label{eq:codebook}
G\sim U(-\frac{1}{N_e}, \frac{1}{N_e})
\end{equation}
where $N_e$ is the size of codebook. 
The overall objective function is followed \cite{vqvae2} and described in the equation (\ref{eq:vae_loss}).
\begin{equation} \label{eq:vae_loss}
\centering
\begin{aligned}[c]
 \mathcal{L}(\mathrm{x},D(\mathrm{e}))= \mathcal{L}_{re} + \mathcal{L}_{vq} 
 \end{aligned}
% \mathcal{L}(\mathrm{x},D(\mathrm(e)=\sum_i^N\left(||sg(f_i)-c_i||+\beta\times||f_i-sg(c_i)|| \right)
\end{equation} where $\mathcal{L}_{re}$ and $\mathcal{L}_{vq}$ are the reconstruction loss and vector quantization loss which are defined in equations (\ref{eq:re_loss}) and (\ref{eq:vq_loss}), respectively.
\begin{equation} \label{eq:re_loss}
    \mathcal{L}_{re} = \|\mathrm{x}-D(\mathrm{e})\|^2_2
\end{equation}
\begin{equation} \label{eq:vq_loss}
    \mathcal{L}_{vq} = \|sg[E(\mathrm{x})]-\mathrm{e}\|_2^2 + \beta\|sg[\mathrm{e}] - E(\mathrm{x})\|^2_2
\end{equation}
where $sg$ represents the stop gradient term. This term is initially set as the identity during forward computation and its partial derivatives are set to zero. Additionally, $\beta$ is a hyperparameter used for tuning. It affects the reluctance to modify the code associated with the encoder output.

\subsection{Similarity score} \label{sub:sim_score}
\begin{figure}[ht]
    \centering
    \includegraphics[width=0.35\textwidth]{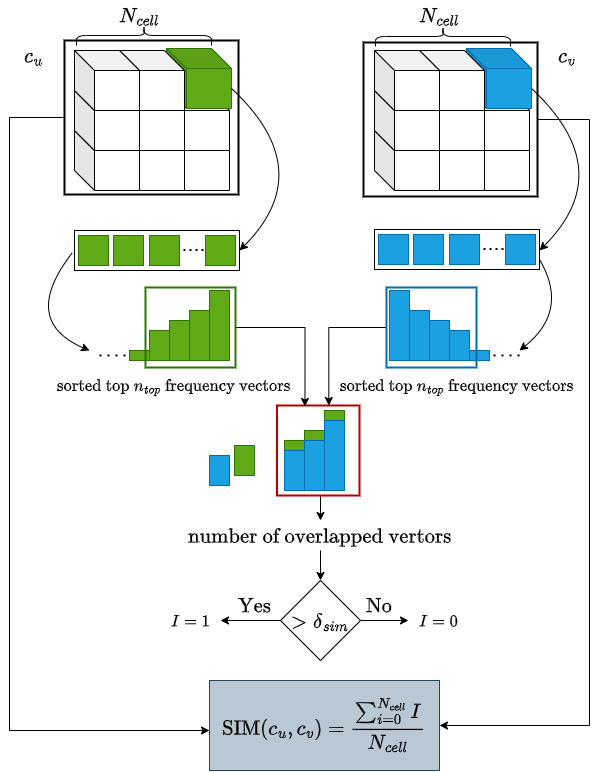}
    \caption{The diagram of similar estimation for two feature maps $c_u$ and $c_v$. $I$ is denoted for the indicator function, which two cells are similar ($I=1$) or not ($I=0$).}
    \label{fig:sim_cal}
\end{figure}

Figure \ref{fig:sim_cal} depicts the procedure for calculating the similarity score between two embedding feature maps, $c_u$ and $c_v$. During this procedure, we initially partition each feature map into a grid consisting of cells of identical size, with $N_{cell}$ cells in total. Our primary idea for computing the similarity score between two feature maps involves partitioning the feature maps into a grid and then comparing the similarity of the associated cells in each grid. Intuitively, if two grids contain a greater number of pairs of comparable cells, their similarity score will be higher.
We iterate through a loop to calculate the similarity between two cells that have the same index in two grids. To begin with, we perform a process of flattening each cell feature, resulting in two sequences of discrete vectors. Next, we group the distinct discrete vector with its corresponding frequency and arrange them in descending order based on frequency. In the subsequent stage, we choose an equal number of top $n_{top}$ frequency distinct vectors from both sides and ascertain the count of overlapping distinct vectors between them. In the final stage, we assess the similarity between two cells by comparing the number of overlapping discrete vectors with a predetermined threshold, denoted as $\delta_{sim}$. Finally, we compute the similarity score between two feature maps, $c_u$ and $c_v$, by dividing the number of comparable cells by the total number of cells in the grid, $N_{cell}$.
% Ultimately, we establish the similarity score between two feature maps, $c_u$ and $c_v$, by calculating the ratio of the number of cells that are comparable to the total number of cells $N_{cell}$ in the grid.
%-----------------------------------
\section{Experiment} \label{sec:exp}
\subsection{Experiment Setting}
\textbf{Datasets.}
We utilize a wide variety of data sets in our experiment. Firstly, we select a dataset to train our projection model. We create a training dataset by combining information from multiple sources. We have selected a cumulative sum of 10,000 images from three publicly available datasets: 3,000 images from the Singapore Maritime dataset \cite{singapore}, 3,000 images from the MODD2 dataset \cite{modd2}, and 4,000 images from the Seaships dataset \cite{seaship}. To ensure an unbiased evaluation, we choose an alternative dataset to examine our system. We selected 35 videos from the RoboWhaler \cite{robowhaler}. These videos are centered and have different durations. The total duration adds up to around 66.37 minutes, with each video being recorded at a frame rate of 12 frames per second. Therefore, a grand total of 47,681 frames were successfully retrieved. We categorize all of these videos by annotating each part into two unique categories: changed and not changed scenes. Out of the total number of frames, specifically 11,369 frames were found to be not scene changed, while the rest 36,312 frames were found to have scene changed.  
% We employ a diverse range of data sets in our experiment. To begin, we choose a dataset to train our projection model. We develop a training set derived from several sources. We have chosen a total of 10,000 photos from three public datasets: 3,000 images from the Singapore Maritime \cite{singapore}, 3,000 images from the MODD2 \cite{modd2}, and 4,000 images from the Seaships \cite{seaship}. In order to guarantee an impartial assessment, we opt for a different dataset to evaluate our system. From the RoboWhaler data set \cite{robowhaler}, we choose 35 camera videos that are centered and have varying duration. The cumulative time amounts to approximately 66.37 minutes, with each video being shot at a frame rate of 12 frames per second. Consequently, a total of 47,681 frames were recovered. We annotate all of these films by categorizing each section into two distinct categories: altered and unaltered. A total of 11,369 frames were identified as not modified, whereas 36,312 remaining frames were identified as changed. 

\textbf{Setting}
We do resizing of the image to the dimensions of $960\times 600 \times 3$, while adding padding to match the width, height, and number of channels. The normalization of all photos is performed using a Gaussian distribution with a mean and standard deviation of 0.5. We set up two hierarchical layers for projection model. In the codebook of the VQ-VAE-2 projection model, we establish that the number of items, denoted as $N_e$, is 512. Every item in the code book consists of 64 discrete vectors. We use the feature map from \textit{bottom} hierarchical with shape of the embedding feature is $150\times 240 \times 64$. We implemented a dual-layer structure for the encoder network of the VQ-VAE-2 to enhance its efficiency and reduce its computational burden. The VQ-VAE-2 model contains a total of 1.3 million trainable parameters. The PyTorch library was used to create the model, which was trained for 100 epochs using the Adam optimizer \cite{adam} with a learning rate of $3e^-4$. 

% To optimize the evaluation process and reduce inference time, we implement frame skipping with $s$ is 4. This means that we only consider every 4th frame, resulting in a computation rate of 3 frames per second. We do not utilize a stride window in order to optimize computing efficiency. 
We use frame skipping with $s=4$ to optimize the evaluation process and reduce inference time. This means that we only consider every fourth frame, yielding a computation rate of three frames per second. To maximize computing efficiency, we do not use a stride window.
We have established that the duration of each window is 10 seconds, which corresponds to $l=120$ frames. 
To calculate the similarity score, we setup $N_{cell}$ is 25 cells, $\delta_{sim}$ is 2, and $n_{top}$ is 5. We transform the feature map into a grid with dimensions of $5\times 5$. Each cell in the grid has dimensions of $30\times 48\times 64$. To calculate the similarity of grid-level, we specify that the number of most frequently occurring discrete vectors is 10, and the criterion for the top similarity of scalar vectors is 5. The change detection component utilizes the widely-used clustering algorithm K-Means \cite{kmeans} with a fixed number of clusters, specifically 2 clusters representing the categories of changed scene and unchanged scene.  
The execution of all steps is performed on the NVIDIA GeForce RTX 2080 Ti device, which has a capacity of 11 GB.  

\textbf{Evaluation metrics.} We do applying the precision, recall and F1-score for evaluating scene change detection. 

\subsection{Results}
\begin{figure}[ht]
    \centering
    \includegraphics[width=0.35\textwidth]{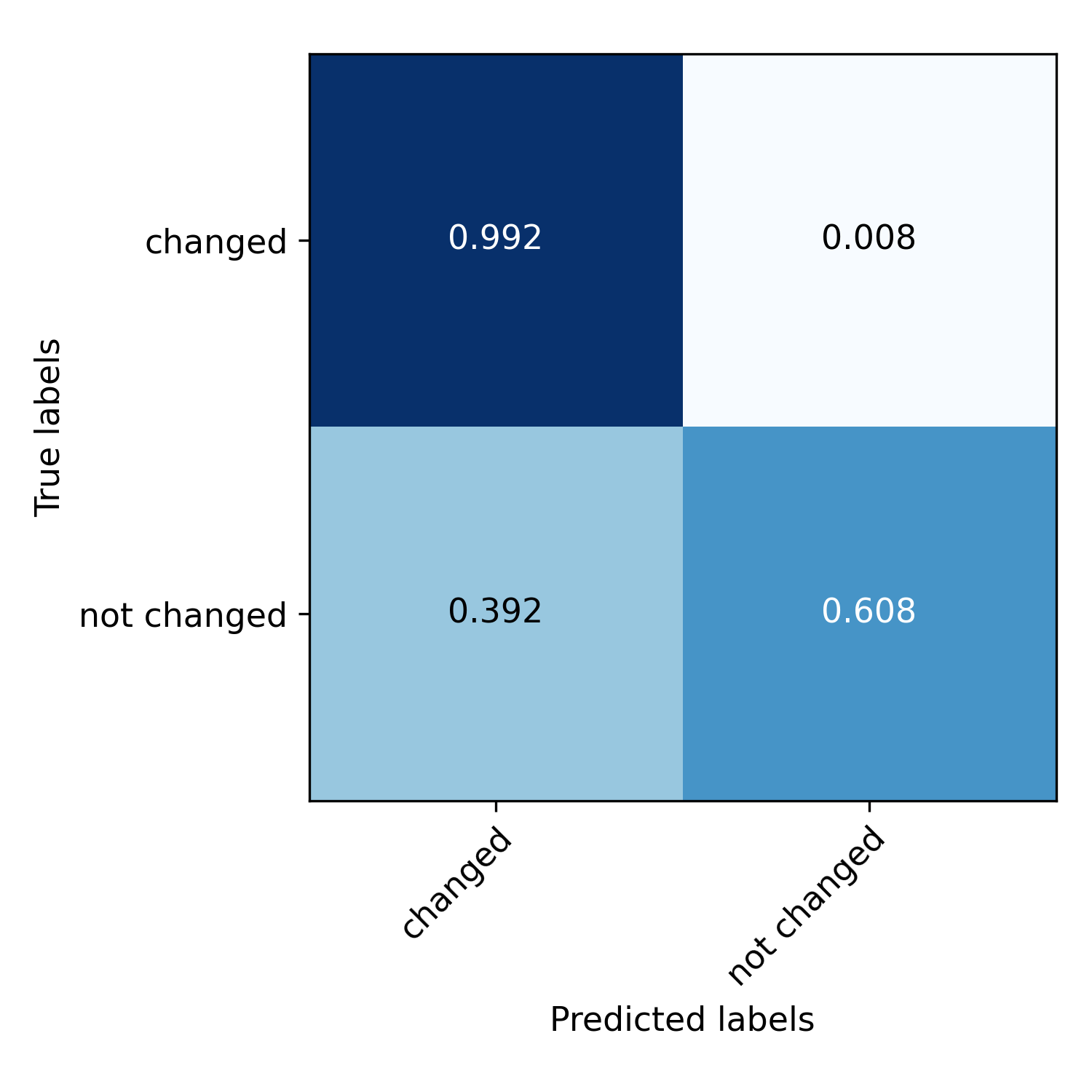}
    \caption{The confusion matrix results of {\OurName}.}
    \label{fig:result_cm}
\end{figure}

\textbf{Quantitative result.}
The Figure \ref{fig:result_cm} and Table \ref{table:result_cr} display the comprehensive performance analysis of our {\OurName} technique. The results demonstrate that {\OurName} achieves a high level of accuracy in detecting dynamic scene changes, with a recall rate of 99\%. However, the performance of detecting unchanged scenes is inferior. There are numerous transitional scenes that occur between a distinct, unaltered scene and a modified one that are unclear and challenging to identify.   Adjusting the stride value during the sliding window process might enhance the accuracy in identifying scene boundaries. However, this adjustment may result in increased processing expenses during inference. This is the compromise between the expense of computing and the level of precision. 

The classification report in the Table \ref{table:result_cr} shows model's performance in distinguishing between scene changed and not scene changed classes. The model's precision, recall, and F1-score show differential performance between the two classes. For scene changed, the model has robust results with precision of 0.89, recall of 0.99, and F1-score of 0.94. However, it struggles with not scene changed, with a lower precision of 0.96 but a lower recall of 0.61. The overall accuracy of 0.9 across all samples supports the model's ability to accurately predict both classes.
\begin{table}[ht]
    \centering
    \resizebox{\columnwidth}{!}{%
        \begin{tabular}{lccrc}
        \toprule
                         & Precision & Recall & F1-score & \# Samples \\
        \midrule \hline 
        changed          & 0.89      & 0.99   & 0.94     & 36,312      \\
        cot changed      & 0.96      & 0.61   & 0.74     & 11,369      \\
        \midrule                                         
        Accuracy         & \multicolumn{3}{r}{0.9}       & 47,681      \\
        Macro average    & 0.92      & 0.8    & 0.84     & 47,681      \\
        Weighted average & 0.91      & 0.9    & 0.89     & 47,681      \\
        \bottomrule
        \end{tabular}
    }
\caption{The classification report results of {\OurName}.}
\label{table:result_cr}
\end{table}

\begin{figure*}[ht]
    \centering
    \begin{subfigure}{.9\textwidth}
        \centering
        \includegraphics[width=\columnwidth]{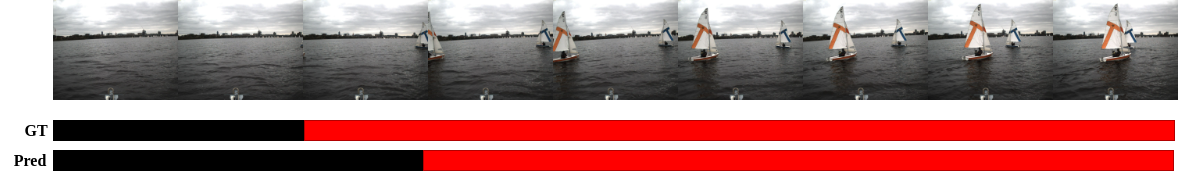}  
        \caption{Example on a part of video prodromos\_2021\_10\_29\_sailboats\_busy.}
        \label{subfig:qualitative_ex1}
    \end{subfigure}
    \begin{subfigure}{.9\textwidth}
        \centering
        \includegraphics[width=\columnwidth]{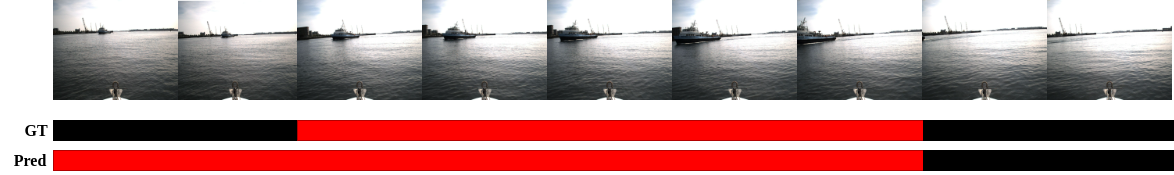}  
        \caption{Example on a part of video philos\_2020\_09\_24\_Gateway\_Enveavor.}
        \label{subfig:qualitative_ex2}
    \end{subfigure}
    \caption{Example of qualitative results of {\OurName} on sampled data of RoboWhaler \cite{robowhaler}. In these example, images in video here are sampled per 1.5 seconds. \protect\tikz \protect\node [rectangle,draw, fill = black!70] at (2.5,-2) {};: not changed scene, \protect\tikz \protect\node [rectangle,draw, fill = red!70] at (2.5,-2) {};: changed scene.}
    \label{fig:qualitative}
\end{figure*}

\textbf{Qualitative analysis.}
We randomly selected a few videos from the RoboWhaler dataset for qualitative analysis. The Figure \ref{fig:qualitative} displays the qualitative analysis of two sampled videos. In this case, the photos were captured at regular intervals of 1.5 seconds from a specific portion of the film in order to clearly observe the transition between scenes. We additionally generate a graph illustrating the correlation between the predicted values (Pred) and the annotated values (GT).  
The Figure \ref{fig:qualitative} show 
In Figure \ref{subfig:qualitative_ex1}, the ground truth (GT) indicates that the scene did not change until the sailboat appeared closely in front of the USVs. However, the prediction incorrectly identifies a small section in the boundary of the modified scene as not being scene changed. The remaining predictions in the sequence accurately correspond to the ground truth. 
In a separate film, Figure \ref{subfig:qualitative_ex2} depicts a large ship in the distance moving at a slow pace towards the USV. The scenario remains unchanged throughout. However, our approach has inaccuracies in identifying the unchanged scene. In the subsequent portion of the video, the model accurately predicts alignment with the annotation. These instances of failure demonstrate that accurately identifying dynamic scene changes in tough scenarios remains a difficult task for implementing practical nautical vision applications.
%-------------------------------------
\section{Conclusion} \label{sec:conclude}
This paper introduces our SeaDSC method, a framework for detecting dynamic scene changes in videos captured by USVs. Our framework consists of three primary components: a feature extraction that aims to project an image into an embedding feature, a similarity scoring component for calculating the magnitude of similarity inside video segments, and a clustering module that groups similar magnitudes into either scene changed or not scene changed segments. As a crucial element of our system, we provide a novel method for calculating similar magnitudes. Our method propose grid similarity calculation that relies on quantized discrete vectors. Our clustering process concludes with the utilization of the basic K-means algorithm. The experimental results on the maritime video dataset RoboWhaler with our annotated data demonstrate the efficacy of our approaches in terms of both accuracy and processing time, making them highly promising for real-world maritime applications. 
Furthermore, this work is the inaugural expansion into marine vision based on our current understanding. Our future work will involve expanding into another area of scene change detection, specifically semantic scene change detection for USVs. 

\section{Acknowledgments}
\label{section:ack}
The work was carried out in the framework of project INNO2MARE - Strengthening the Capacity for Excellence of Slovenian and Croatian Innovation Ecosystems to Support the Digital and Green Transitions of Maritime Regions  (Funded by the European Union under the Horizon Europe Grant N°101087348).
{\small
\bibliographystyle{ieee_fullname}
\bibliography{main}
}

\end{document}